\documentclass[10pt,twocolumn,letterpaper]{article}

\usepackage[pagenumbers]{cvpr} 

\definecolor{cvprblue}{rgb}{0.21,0.49,0.74}
\usepackage[pagebackref,breaklinks,colorlinks,allcolors=cvprblue]{hyperref}
\usepackage{subcaption}
\usepackage{multirow}
\usepackage{utfsym}

\usepackage{listings}
\usepackage{xcolor}
\usepackage{algorithm} 

\definecolor{pybg}{RGB}{255,255,255}
\definecolor{pycomment}{RGB}{34,139,34}   
\definecolor{pykeyword}{RGB}{178,34,34}   
\definecolor{pyname}{RGB}{10,36,106}      

\lstdefinestyle{algo1-python}{
  language=Python,
  basicstyle=\ttfamily\scriptsize,
  keywordstyle=\color{pykeyword}\bfseries,
  commentstyle=\color{pycomment},
  stringstyle=\color{pyname},
  mathescape=true,
  frame=none, framerule=0pt, rulesep=0pt, xleftmargin=0pt, xrightmargin=0pt,
  showstringspaces=false, tabsize=2, columns=fullflexible, keepspaces=true,
  breaklines=true, breakatwhitespace=false, 
}

\def\paperID{7350} 
\def\confName{CVPR}
\def\confYear{2026}

\title{\textit{OpenUS}: A Fully Open-Source Foundation Model for Ultrasound Image Analysis via Self-Adaptive Masked Contrastive Learning}

\author{
Xiaoyu Zheng\textsuperscript{1,*} \quad
Xu Chen\textsuperscript{1} \quad
Awais Rauf\textsuperscript{1} \quad
Qifan Fu\textsuperscript{1}\\[2pt]
Benedetta Monosi\textsuperscript{2} \quad
Felice Rivellese\textsuperscript{2} \quad
Myles J. Lewis\textsuperscript{2}\\[2pt]
Shaogang Gong\textsuperscript{3} \quad
Gregory Slabaugh\textsuperscript{1,*}\\[6pt]
\textsuperscript{1}Digital Environment Research Institute (DERI), 
Queen Mary University of London\\
\textsuperscript{2}The William Harvey Research Institute, 
Queen Mary University of London\\
\textsuperscript{3}School of Electronic Engineering and Computer Science, 
Queen Mary University of London\\
London, UK\\[3pt]
\textsuperscript{*}{\tt\small zhengxiaoyu0427@gmail.com/g.slabaugh@qmul.ac.uk}
}

\begin{document}
\maketitle
\begin{abstract}
Ultrasound (US) is one of the most widely used medical imaging modalities, thanks to its low cost, portability, real-time feedback, and absence of ionizing radiation. However, US image interpretation remains highly operator-dependent and varies significantly across anatomical regions, acquisition protocols, and device types. These variations, along with unique challenges such as speckle, low contrast, and limited standardized annotations, hinder the development of generalizable, label-efficient ultrasound AI models. In this paper, we propose \textit{OpenUS}, the first reproducible, open-source ultrasound foundation model built on a large collection of public data.  \textit{OpenUS} employs a vision Mamba backbone, capturing both local and global long-range dependencies across the image.  To extract rich features during pre-training, we introduce a novel self-adaptive masking framework that combines contrastive learning with masked image modeling. This strategy integrates the teacher's attention map with student reconstruction loss, adaptively refining clinically-relevant masking to enhance pre-training effectiveness. \textit{OpenUS} also applies a dynamic learning schedule to progressively adjust the difficulty of the pre-training process. To develop the foundation model, we compile the largest to-date public ultrasound dataset comprising over 308K images from 42 publicly available datasets, covering diverse anatomical regions, institutions, imaging devices, and disease types. Our pre-trained \textit{OpenUS} model can be easily adapted to specific downstream tasks by serving as a backbone for label-efficient fine-tuning. Code is available at \url{https://github.com/XZheng0427/OpenUS}.
\end{abstract}    
\begin{figure}[t]
\centering
\includegraphics[width=0.48\textwidth]{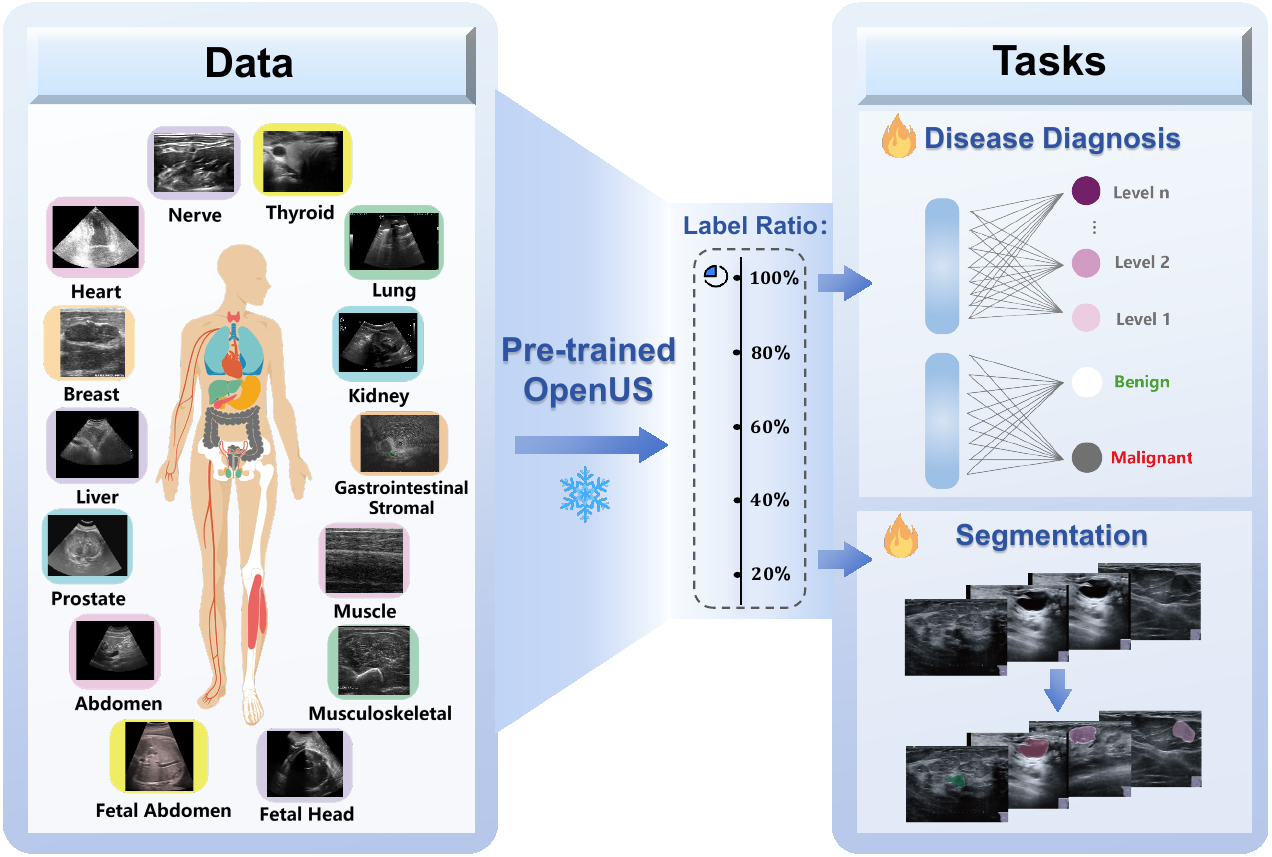} 
\caption{Overview of Universal US Foundation Model.}
\label{fig0}
\end{figure}

\section{Introduction}
\label{sec:intro}

Ultrasound (US) imaging is widely used in healthcare due to its affordability, portability, real-time feedback, and safety \cite{b1}. With applications ranging from fetal monitoring to cardiac, abdominal, and musculoskeletal imaging, US plays a central role in both diagnostics and image-guided interventions. Recent advances in computer-aided US analysis have enabled automated classification, segmentation, and disease detection \cite{b2,b3,b4}, yet the clinical adoption of supervised deep learning methods remains limited by the scarcity of large, diverse, and well-annotated datasets. Moreover, US images vary considerably across organs, acquisition protocols, and devices, complicating the generalization of task-specific models. These challenges have created growing interest in US foundation models that can learn transferable representations across anatomical regions and clinical tasks, facilitating label-efficient adaptation to new applications \cite{b5}. Such models have the potential to unlock scalable, intelligent US workflows and support broader integration into real-world healthcare systems.

Recently, visual foundation models \cite{b47yuan2021florence} have attracted significant attention in the field of natural image analysis, demonstrating strong potential for developing generalized and robust representations. These models are typically developed using self-supervised pre-training methods on large-scale, unlabeled datasets  \cite{b6, b7, b8}. To effectively learn meaningful representations in self-supervised learning (SSL), researchers have developed various visual pretext tasks \cite{b9,b10,b11}, broadly categorized into contrastive and generative approaches. Motivated by successes in natural image analysis, extending foundation models to US imaging offers promising potential for efficiently developing versatile models applicable across various tasks and anatomical structures. However, transferring foundation models from natural images to US images remains challenging due to fundamental differences in imaging principles \cite{48zhang2024challenges}.  Therefore, there is a critical need to develop a foundation model specifically designed for US imaging, accompanied by comprehensive evaluation of its versatility and adaptability across diverse downstream tasks. However, current US foundation models \cite{b5,b44} rely on proprietary US images for pre-training, hindering reproducibility. Addressing this issue is essential for developing a fully reproducible, open-source US foundation model.

In this paper, we propose \textit{OpenUS}, a foundation model developed for universal US imaging analysis with label efficiency and downstream task adaptability, as shown in Fig.~\ref{fig0}. 
\textit{OpenUS} is built solely on publicly available data, making it the first reproducible ultrasound foundation model.
To achieve self-supervised pre-training on clinically relevant and generalizable US features, we use vision Mamba, based on VMamba \cite{49liu2024vmamba, 67zheng2025xfmamba}, to capture long-range spatial dependencies. Second, we propose a novel self-adaptive masking strategy that combines the teacher's attention map with the student's reconstruction loss. This approach enables the model to more accurately target and mask challenging, clinically relevant regions by using student feedback to correct for potential inaccuracies in the teacher's attention. Additionally, we incorporate multi-view masked modeling into the contrastive learning framework, which not only encourages the model to learn more discriminative representations but also compels it to capture more refined, pixel-level features. Extensive experiments confirm that \textit{OpenUS} exhibits strong generalizability, superior performance, and enhances label efficiency for downstream task fine-tuning. Overall, our contributions are summarized as follows:
\begin{itemize}
    \item We are the first to develop a fully reproducible, open-source US foundation model, \textit{OpenUS}.
    \footnote{Concurrent to this work, Zhang et al.\ introduced \textit{EchoCare}~\cite{zhang2025echocare}. At the time of writing, this EchoCare appears on ArXiv, and its code and pretrained weights have not been released.}   
    \textit{OpenUS} is designed as a plug-and-play module to improve the label-efficiency and performance of various downstream US image analysis tasks. \textit{OpenUS} is trained on large-scale, fully open-access datasets spanning diverse organs, institutions, and devices.
    \item We propose a multi-view masked contrastive representation learning framework that integrates global-local-view masked modeling with a contrastive approach for pre-training. This framework effectively overcomes the limitations of conventional contrastive methods in capturing dense pixel-level dependencies by enabling the prediction of pixel intensities within masked regions.
    \item We introduce a novel self-adaptive masking strategy to extract clinically relevant and generalizable US features.
    Teacher attention-guided masking and student reconstruction-loss masking synergistically enhance the robust extraction of effective US image features.
    \item We conduct extensive experiments on 38 fully public datasets for pre-training and 4 datasets for classification and segmentation downstream tasks to evaluate the performance of \textit{OpenUS} compared to existing methods.
\end{itemize}

\begin{figure*}[t]
\centering
\includegraphics[width=1\textwidth]{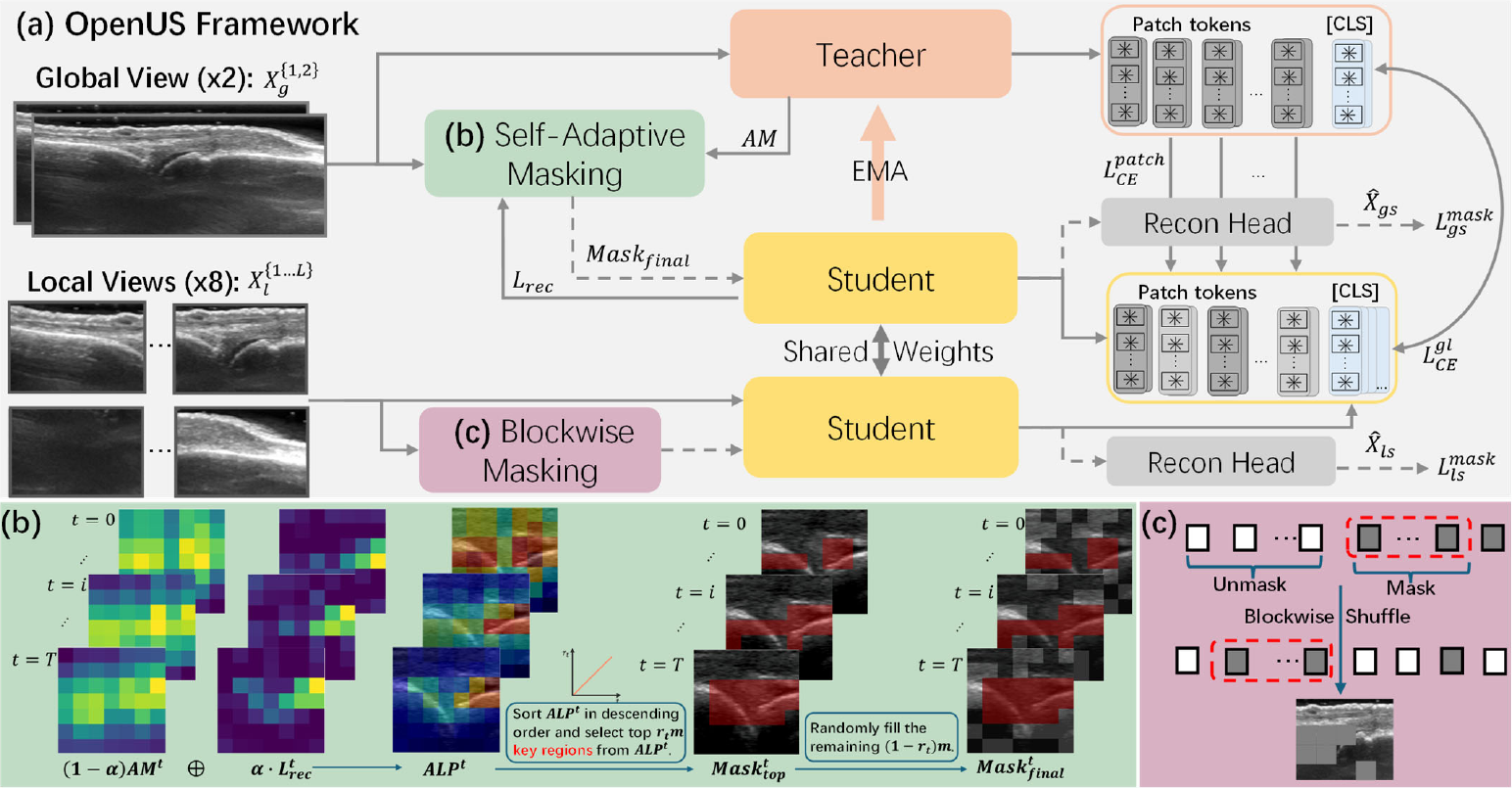} 
\caption{(a) The \textit{OpenUS} pipeline including the Teacher and Student models, masking approaches and reconstructions heads. For global and local views, we design two distinct masking strategies: (b) self-adaptive masking and (c) random block-wise masking.  Both are integrated with masked image reconstruction and contrastive learning.}
\label{fig1}
\end{figure*}

\section{Related Work}
\label{sec:relatedwork}

\subsection{Contrastive Learning}
SSL adopts unlabeled data to extract meaningful representations without the need for explicit supervision, leading to considerable progress in pattern recognition \cite{b29,b30}. In this paradigm, contrastive learning (CL) initially attracted significant interest in both computer vision \cite{b31,b32,51caron2021emerging} and medical imaging domains \cite{b33,b34}. CL captures rich semantic representations by bringing similar image pairs closer together and pushing dissimilar pairs further apart in the feature domain. However, CL depends on specifically designed data augmentation techniques to generate paired images for learning \cite{b35}. In applications that demand accurate pixel-level embeddings for downstream tasks, i.e., classification, segmentation and detection, CL approaches may overlook the subtle details essential for identifying small organs or lesion regions as they tend to emphasize the broad, discriminative features rather than focusing on the fine-grained features.

\subsection{Masked Image Modeling}
Inspired by the breakthroughs achieved with masked language modeling in natural language processing, masked image modeling (MIM) has garnered much attention in computer vision, i.e., the bidirectional encoder representation from image transformers (BEiT) \cite{b19}, masked autoencoder (MAE) \cite{b22} and the simple framework for MIM (SimMIM) \cite{b21}. Both He et al. \cite{b22} and Xie et al. \cite{b21} propose to mask random image patches and then predict the corresponding RGB pixel values. Following the introduction of BEiT and MAE, iBOT \cite{b38} emerged as an online tokenizer adaptation of BEiT designed to overcome its shortcoming of capturing only low-level semantics within local details. While MIM strategies have been successful in natural images, they can overlook the detailed and clinically meaningful patterns in medical images, such as anatomical boundaries or diagnostically relevant regions, which can limit the effectiveness of self-supervised pre-training. Recent advancements in MIM have shifted focus towards novel reconstruction targets \cite{b36,b39} and masking strategies \cite{b26,b27,b28}. Specifically, Liu et al. \cite{b28} proposed an attention-driven masking approach to overcome the limitations of random masking in fully utilizing semantic information. However, novel MIM strategies specifically designed for medical image analysis remain underexplored.

\subsection{Foundation Model for US Image Analysis}
In recent years, foundation models based on SSL have attracted significant attention in medical image analysis, including applications in ultrasound (US) imaging. Specifically, US imaging introduces unique challenges compared to other modalities, including speckle patterns, motion artefacts, and geometric discrepancies across probe types and manufacturers \cite{b46}. USFM \cite{b5} employs a spatial-frequency dual-MIM method to extract effective features from low-quality US images. Kang et al. \cite{b44} proposed deblurring MIM that integrated the US-specific deblurring task into the standard MAE framework. Instead of only reconstructing masked patches, the model also learns to restore the characteristic low signal-to-noise US texture and capture finer detail. Basu et al. \cite{b45} introduced FocusMAE, a region-focused masked modeling approach for US videos aimed at detecting gallbladder carcinoma. However, this method exhibits certain limitations when applied to general purpose US imaging. While MIM has become the dominant paradigm for self-supervised ultrasound pre-training, relying on MIM alone can limit the ability to learn rich, discriminative representations needed for downstream tasks. 

\section{Method}
Fig.~\ref{fig1} illustrates the overall pipeline of \textit{OpenUS}. Here, we first discuss the preliminaries of vision Mamba and self-distillation MIM. Then, we introduce our \textit{OpenUS} pipeline and provide a detailed explanation of our proposed self-adaptive masking strategy.

\begin{figure}[t]
\centering
\includegraphics[width=0.48\textwidth]{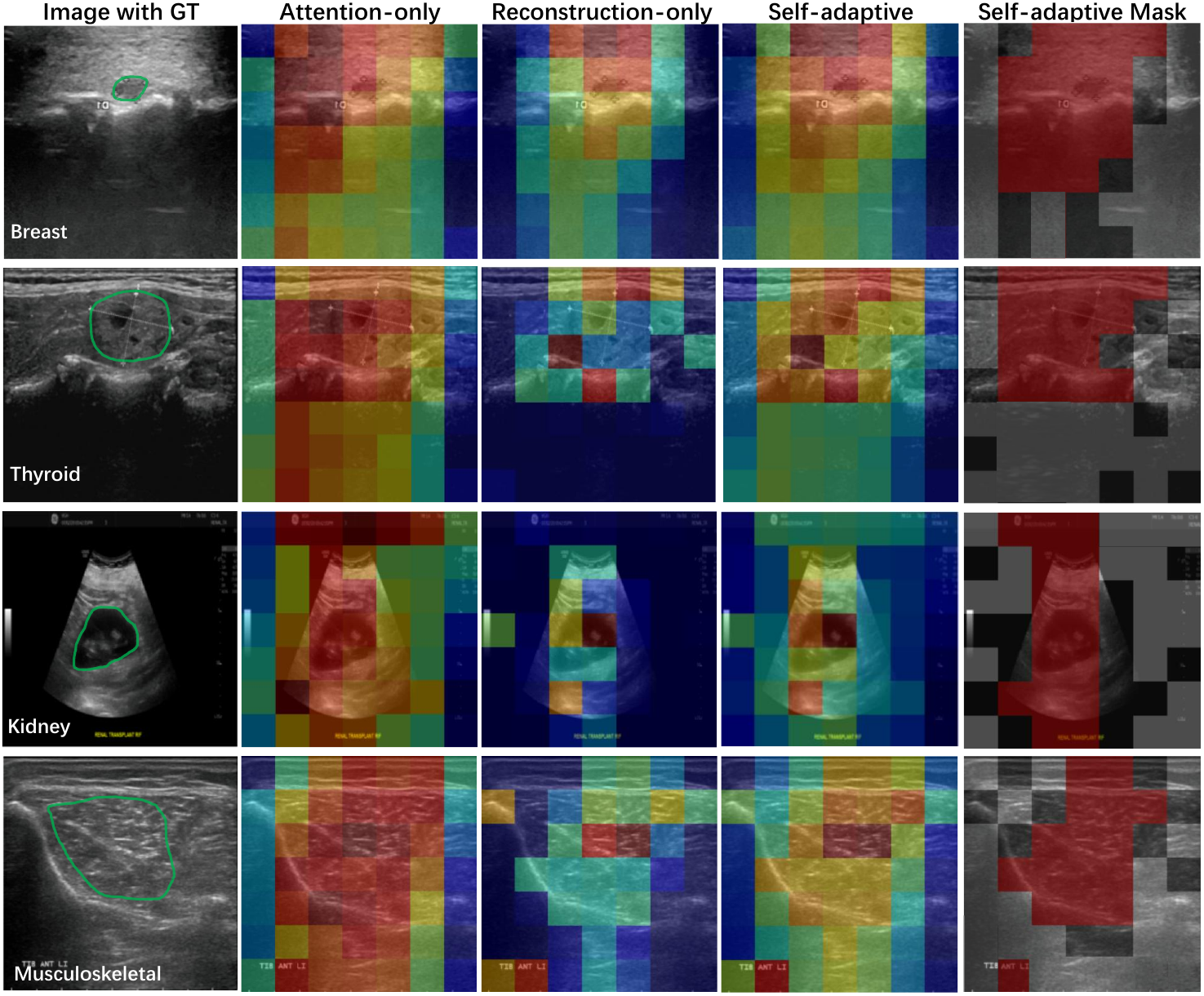} 
\caption{Visual comparison with segmentation ground truths, attention-only, reconstruction loss and self-adaptive $ALP$ scores. In the last column, the red areas tend to overlap with clinically relevant regions, while the dark grey regions represent the remaining randomly masked areas.}
\label{fig2}
\end{figure}

\subsection{Preliminaries}
\subsubsection{Vision Mamba (VMamba).} Here we describe the common backbone VMamba \cite{49liu2024vmamba}, which has a hierarchical encoder design. Specifically, an input image $\mathbf{x} \in \mathbb{R}^{H \times W \times C}$, is first partitioned into patches by a stem module, yielding a 2D feature map with the spatial dimension of $\frac{H}{4} \times \frac{W}{4}$, where $(H,W)$ denotes the height and width of the original image, $C$ is the number of the channels. Hierarchical representations are then created through multiple network stages, achieving resolutions of $\frac{H}{8}\times \frac{W}{8}$, $\frac{H}{16}\times \frac{W}{16}$, and $\frac{H}{32}\times \frac{W}{32}$. Each stage includes a down-sampling layer followed by the Visual State Space (VSS) block that is derived from the zero-order hold (ZOH) method \cite{53gu2023mamba}, 
{\small
\begin{equation}
\label{eq1}
\mathbf{h}_b = e^{\mathbf{A}(\Delta_a + \cdots + \Delta_{b-1})} \left( \mathbf{h}_a + \sum_{i=a}^{b-1} \mathbf{B}_i u_i e^{-\mathbf{A}(\Delta_a + \cdots + \Delta_i)} \Delta_i \right)
\end{equation}
}where $[{\mathbf{a}}, {\mathbf{b}}]$ is the corresponding discrete step interval. $\mathbf{A} \in \mathbb{R}^{N \times N}$, $\mathbf{B} \in \mathbb{R}^{N \times N}$, $\mathbf{h}$ denotes the hidden state, and $\Delta$ is the time-scale parameter.

\subsubsection{Self-distillation masked image modeling.} Self-distillation, which employs a moving average of the student model as a teacher, has been explored for SSL in BYOL \cite{50grill2020bootstrap}. This approach was subsequently extended to vision transformers in DINO \cite{51caron2021emerging}, where the distillation loss is applied globally to the [cls] token. iBOT \cite{b38} addresses this task through MIM, where the loss is computed densely over the masked tokens. Following iBOT, the transformer encoder is succeeded by a head, comprising a Multi-Layer Perceptron (MLP) and scaled softmax, enabling the output token embeddings to be interpreted as probabilities. The teacher parameters $\theta'$ are updated from the student parameters $\theta$ using an exponential moving average (EMA), following the update rule $\theta' \leftarrow \lambda \theta' + (1 - \lambda) \theta$, where $\lambda \in [0,1)$ controls the momentum of the update. For each input image, two augmented global views of standard resolution are generated, resulting in tokenized images $\mathbf{P}^a$, $\mathbf{P}^b$ and corresponding mask vectors $\mathbf{m}^a$, $\mathbf{m}^b$. For each view $v \in \{a, b\}$ and its respective masked token, the MIM objective is to minimize the reconstruction loss between the student's output ($f_\theta$ applied to the masked input $\mathbf{\tilde{P}}^v$) and the teacher's output ($f_{\theta'}$ applied to the non-masked input $\mathbf{P}^v$),
{\small
\begin{equation}
\label{eq2}
\mathcal{L}_{\text{MIM}} = - \sum_{\mathbf{v} \in \mathbf{V}} \sum_{i=1}^{n} m_i^{\mathbf{v}} f_{\theta'}(\mathbf{P}^{\mathbf{v}}_i) \log\big(f_\theta({\mathbf{P}}^{\mathbf{v}}_i)\big)
\end{equation}
}

\subsection{OpenUS via Global-local Masked Contrastive Representation Learning} 
The pipeline of our proposed \textit{OpenUS} is illustrated in Fig.~\ref{fig1}(a). It employs contrastive learning combined with global-local mask modeling to simultaneously learn representations that are both fine-grained and discriminative. The contrastive learning component of our \textit{OpenUS} model follows iBOT \cite{b38}, which also adopts a self-distillation approach to facilitate representation learning. 

Given a US image, two types of views are created under random data augmentation, i.e., global views $\mathbf{X}_g^i$ and local views $\mathbf{X}_l^j$. During pre-training, the global views are fed into both the teacher and student networks, while the local views are only fed into the student network. The network output is normalized using a softmax function with temperature $\tau$ to yield the probability distribution $p = \mathrm{softmax}(f / \tau)$. Furthermore, the cross-entropy loss function is employed to compute the loss $L^{GL}_{CE}$ between the teacher's global-view $[\mathrm{cls}]$ tokens and the student's local-view $[\mathrm{cls}]$ tokens, as well as the loss $L^{patch}_{CE}$ between the student network outputs for the masked-view patch tokens and the teacher network outputs for the non-masked-view patch tokens. Building upon iBOT, we incorporate a global-local view mask reconstruction pre-training task into the contrastive learning framework to enhance dense pixel dependencies, which are critical for dense prediction tasks such as segmentation. Specifically, the reconstruction head leverages the contextual information from unmasked patches to regress the dense pixel intensities of the masked patches. The loss for global view reconstruction is  $\mathcal{L}^{mask}_{gs} = \frac{1}{N^{mask}_{gs}} \sum_{i=1}^{G} \left\| \hat{\mathbf{X}}^{i}_{g} - \mathbf{X}^{i}_{g} \right\|$.

Due to the low contrast between lesions and normal tissues in US images, global views may have difficulty accurately capturing local details. Hence, we employ the random blockwise masking strategy \cite{b19} on local views to capture more fine-grained and detailed information, as shown in Fig.~\ref{fig1}(c). The loss for local view reconstruction is $\mathcal{L}^{mask}_{ls} = \frac{1}{N^{mask}_{ls}} \sum_{i=1}^{L} \left\| \hat{\mathbf{X}}^{i}_{l} - \mathbf{X}^{i}_{l} \right\|$. 

\textit{OpenUS} leverages both a contrastive loss and a reconstruction loss during optimization, with the total loss defined as: $\mathcal{L}_{\text{total}} = \mathcal{L}^{gl}_{\text{CE}} + \mathcal{L}^{patch}_{\text{CE}} + \mathcal{L}^{mask}_{gs} + \mathcal{L}^{mask}_{ls}$. The student network updates its parameters $\theta_s$ by minimizing the total loss $\mathcal{L}_{\text{total}}$, while the teacher network parameters $\theta_t$ are updated using an EMA of the student weights, following the rule: $\theta_t \leftarrow \lambda \theta_t + (1 - \lambda) \theta_s$, where $\lambda$ defines the momentum coefficient. By integrating contrastive learning with MIM, the model learns both holistic discriminative features and fine-grained pixel details, enhancing its capability to process complex, clinically relevant features from US images. 

\subsection{Self-adaptive Masking Strategy}
MIM employs random and attention-only mask strategies to extract meaningful representations from pre-training images. While these strategies are effective for general image datasets with stable and well-curated distributions, they do not adequately address the unique characteristics inherent in medical images. 
US images often exhibit strong speckle, low contrast between tissue boundaries, and appearance variability caused by differences in probe type, acquisition protocol, and operator technique. Furthermore, diagnostically important structures may occupy only small or subtle regions of the image, making them harder to capture with purely random or attention-only masking. Therefore, to address the challenges outlined, we propose a self-adaptive masking strategy utilizing a teacher-student feedback mechanism, as depicted in Fig.~\ref{fig1}(b). The specifics are detailed below.

\subsubsection{Semantic Information Extractor.} Our network employs a self-attention mechanism known as 2D Selective Scan (SS2D) \cite{49liu2024vmamba}, which boosts the learning capacity of the network. Specifically, we take the US image, from which we sample the global views $\{\mathbf{X}^{i}_g \in \mathbb{R}^{3 \times H_g \times W_g} \}_{i=1}^{G}$. Each image is then divided into $N=H_gW_g/P^2$ patches, which are mapped into patch tokens and fed into the multiple teacher VSS blocks with the down-sampling layer to create hierarchical representations. The network incorporates the learnable class token and patch tokens, which represent the global and local features extracted by the network across the spatial dimension. Unlike Vision Transformers (ViTs) that incorporate a learnable class token ($\mathrm{cls}$), our network utilizes average pooling ($avgpool$) to derive the class token from the patch tokens. Given the intermediate token $\mathbf{x}^n \in \mathbb{R}^{(N+1) \times D} $ from block $n$, the token for the subsequent block is computed as follows: 
{\small
\begin{equation}
\label{eq3}
\left\{
\begin{aligned}
\mathbf{x}^{n} &= SS2D \left(LN(\mathbf{x}^{(n-1)}) \right) + \mathbf{x}^{n-1}, \\
\mathbf{x}^{n+1} &= MLP \left(LN(\mathbf{x}^{n}) \right) + \mathbf{x}^{n}
\end{aligned}
\right.
\end{equation}
}where $SS2D$, $LN$ and $MLP$ represent the 2D selective scan attention, layer normalization and multi-layer perception, respectively. We can obtain the attention map of the last layer of the VSS blocks by computing the correlation between the query embeddings $Q$ and the key embeddings $K$, a mechanism similar to self-attention of ViT. 
Specifically, based on Eq.~\ref{eq1}, the discretized solution of time-varying SSMs can be defined as:
{\small
\begin{equation}
\label{eq4}
\mathbf{h}_b = \mathbf{w}_T \odot \mathbf{h}_a + \sum_{i=1}^{T} \frac{\mathbf{w}_T}{\mathbf{w}_i} \odot \left(\mathbf{K}_i^\top \mathbf{V}_i \right)
\end{equation}
}where $\odot$ represents the element-wise product between matrices. 
Building upon the hidden state $\mathbf{h_b}$ (Eq.~\ref{eq4}), the output of SSM, i.e., $\mathbf{Y}^{(i)}$, can be defined as:
{\small
\begin{equation}
\label{eq5}
\mathbf{Y}^{(i)} = \left[\mathbf{Q} \odot \mathbf{w}^{(i)} \right] \mathbf{h}_a^{(i)} 
+ \left[ \left(\mathbf{Q} \odot \mathbf{w}^{(i)} \right) \left( \frac{\mathbf{K}}{\mathbf{w}^{(i)}} \right)^\top \odot \mathbf{M} \right] \mathbf{V}^{(i)}
\end{equation}
}where $\mathbf{h_a} \in \mathbb{R}^{(D_k)}$ is the hidden state at step $a$, $\mathbf{M}$ defines the temporal mask matrix. Therefore, the attention map ($AM$) is calculated by $\mathbf{Q} \mathbf{K}^\top$ and $\left( \mathbf{Q} \odot \mathbf{w} \right) \left( \frac{\mathbf{K}}{\mathbf{w}} \right)^\top$, which is capable of obtaining an approximation of the clinically relevant lesion region of the US image.

\subsubsection{Reconstruction Loss Predictor.} Our network incorporates a reconstruction head employing a lightweight decoder composed of a few linear layers to reconstruct the masked regions. Concretely, let $\mathbf{Mask}^t_{final}$ denote the newly generated mask, the reconstruction loss can be formulated as:
{\small
\begin{equation}
\label{eq6}
\mathcal{L}^t_{\text{rec}} = \mathcal{M}\left(d_{s}(f_{s}(\mathbf{x} \odot \mathbf{Mask}^t_{\text{final}})),\, \mathbf{x} \odot (1 - \mathbf{Mask}^t_{\text{final}})\right)
\end{equation}
}where $\odot$ denotes element-wise dot product, and $\mathbf{x} \odot {(1-\mathbf{Mask}^t_{final})}$ represents masked (invisible) patches and vice versa. $\mathcal{M}(\,\cdot\,,\,\cdot\,)$ represents the similarity measurement, i.e., $l_2$-distance. We utilize the EMA of the student's reconstruction loss ($\mathcal{L}^t_{\text{rec}}$), which provides a more stable loss map highlighting regions that consistently pose challenges for the student. Furthermore, we perform an \textit{argsort} operation on $\mathcal{L}^t_{\text{rec}}$ to rank the masked regions based on their reconstruction losses in descending order. This ranking directs the student network's attention toward reconstructing the most difficult regions, thereby effectively predicting clinically relevant features.

\subsubsection{Mask Generator: Teacher-Student Feedback.} An attention-only approach can become repetitive, repeatedly masking regions the student already finds easy, while a reconstruction-only strategy risks focusing on noisy or irrelevant patches that are difficult but not informative. Our method avoids these pitfalls by integrating the teacher's ``top-down'' attention to critical information with the student's ``bottom-up'' feedback regarding challenging aspects. This synergy ensures the model consistently targets the most valuable and important regions for feature learning, accelerating convergence and leading to more robust representations. 

Specifically, the first is the teacher attention map ($\mathbf{AM}_t$), where higher values indicate regions considered important by the teacher network. The second is the student reconstruction loss map ($\mathbf{L}_{rec_s}$), where higher values indicate patches the student finds challenging to reconstruct. These two inputs are integrated into a unified metric, which we term the \textit{Adaptive Learning Priority} ($\mathbf{ALP}$) score. For each image patch, the $\mathbf{ALP}$ score is calculated as a weighted sum:
{\small
\begin{equation}
\label{eq7}
\mathbf{ALP} = (1-\alpha) \cdot \mathbf{AM}_t + \alpha \cdot \mathbf{L}_{\text{rec}_s}
\end{equation}
}Here, the hyper-parameter $\alpha \in [0,1]$  is a crucial balancing factor that we dynamically schedule to adapt to the model's learning stage. Following a cosine decay schedule, $\alpha$ gradually increases from a low value at the beginning of training to a higher value towards the end. This curriculum ensures that the masking strategy initially prioritizes the stable guidance from the teacher and later transitions to emphasize the more nuanced difficulty feedback from the student, creating a more robust and efficient training process. 

Additionally, in the initial training stages, a high $ALP$ score may not always correlate with clinically relevant importance, as the model is still learning to effectively target the informative regions. To this end, we also adopted a dynamic easy-to-hard mask generation strategy \cite{b26,54li2024anatomask}, wherein the proportion of important regions within the masked areas gradually increases, thereby providing progressive guidance that directs the model's $ALP$ score incrementally toward important regions. As presented in Fig.~\ref{fig1}(b), for each training epoch $t$, $r_t$ of the mask patches are generated by $\mathbf{ALP}^t$, and the remaining $1-r^t$ of the masked regions in $\mathbf{Mask}_{final}^t$ are randomly selected. Specifically, $r_t=r_0+\frac{t}{T}(r_t-r_0)$, where $T$ is the total training epochs, and $r_0,r_t \in [0,1]$ are two tunable hyper-parameters which we empirically set to 0.1 and 0.9, respectively. As demonstrated in Fig.~\ref{fig2}, we identify the clinically relevant regions often demonstrate high $ALP$ scores, indicative of rich semantic information for the network to learn. 

\begin{algorithm}[t]
\caption{Pseudo-Code of SAM in a PyTorch-like Style.}
\label{alg1}
\captionsetup{width=\columnwidth} 

\begin{minipage}{\columnwidth} 
\begin{lstlisting}[style=algo1-python, linewidth=\columnwidth]
# m_s, m_t: networks for student and teacher
# rat_m: masking ratio; thr_m: masking threshold
# mask_pre: mask from previous epoch output

# teacher attention
t_out, AM = model_t.Mamba(x)
# student reconstruction loss
s_out, rec_x = model_s.Mamba.recon_head(x * mask_pre)
L_rec  = (rec_x - x[~mask_pre]) ** 2
mask = sam_gen(AM, L_rec, rat_m, thr_m)

# self-adaptive masking (SAM) generator
def sma_gen(AM, L_rec, rat_m, thr_m):
    # compute ALP score
    ALP = (1 - $\alpha$) $\cdot$ AM + $\alpha$ $\cdot$ L_rec
    N = ALP.size(1)
    n_mask_patches = ceil(N * rat_m)
    top_k_mask = floor(N * thr_m)
    topIdx = topk_indices(ALP, top_k_mask) 
    n = n_mask_patches - top_k_mask
    if n > 0:
        nonTop = all_indices(N) \ topIdx
        addIdx = random_sample(nonTop, n)
        idx = concat(topIdx, addIdx)
    else:
        idx = topIdx[:, :n_mask_patches]
    mask = zeros(B, N); mask[B, idx] = 1
    
    return mask
\end{lstlisting}
\end{minipage}
\end{algorithm}

\begin{figure*}[ht]
  \centering
  \begin{minipage}[t]{0.485\textwidth}
    \centering
    \captionof{table}{Quantitative results on classification tasks.}
    \scriptsize
    \setlength{\tabcolsep}{2.5pt}
    \begin{tabular}{c c c c c c}
      \toprule
      Model           & Pretraining & \multicolumn{2}{c}{Fetal Planes}      & \multicolumn{2}{c}{BUSI}             \\
      \cmidrule(lr){3-4} \cmidrule(lr){5-6}
                                &             & ACC(\%)       & F1(\%)        & ACC(\%)        & F1(\%)         \\
      \midrule
      \multicolumn{6}{c}{Supervised}\\
      \midrule
                  ResNet50       & ImageNet    & 82.6 $\pm$ 0.3 & 79.5 $\pm$ 0.5 & 76.3 $\pm$ 0.6 & 77.1 $\pm$ 0.4 \\
                  ViT             & ImageNet    & 80.3 $\pm$ 0.7 & 75.4 $\pm$ 0.8 & 69.8 $\pm$ 0.4 & 49.2 $\pm$ 0.7 \\
                  VMamba          & ImageNet    & 85.6 $\pm$ 0.5 & 84.6 $\pm$ 0.3 & 83.7 $\pm$ 0.5 & 82.8 $\pm$ 0.3 \\
      \midrule
      \multicolumn{6}{c}{SSL/FM}\\
      \midrule
                  SimMIM          & ImageNet    & 87.7 $\pm$ 0.8 & 86.1 $\pm$ 0.6 & 80.2 $\pm$ 0.3 & 78.3 $\pm$ 0.4 \\
                  DINO+ViT        & ImageNet    & 87.8 $\pm$ 0.6 & 85.3 $\pm$ 0.7 & 82.3 $\pm$ 0.5 & 81.7 $\pm$ 0.7 \\
                  DINOv2+ViT      & ImageNet    & 88.4 $\pm$ 0.3 & 86.1 $\pm$ 0.4 & 85.2 $\pm$ 0.4 & 83.5 $\pm$ 0.5 \\
                  iBOT+ViT        & ImageNet    & 88.3 $\pm$ 0.2 & 86.5 $\pm$ 0.5 & 85.1 $\pm$ 0.6 & 83.3 $\pm$ 0.7 \\
                  DeblurringMAE   & US-Esaote-280K & 88.2 $\pm$ 0.4 & 85.9 $\pm$ 0.2 & 84.9 $\pm$ 0.7 & 82.4 $\pm$ 0.3 \\
                  USFM            & 3M-US-2M      & \underline{90.2 $\pm$ 0.5} & \underline{89.1 $\pm$ 0.6} & \underline{87.5 $\pm$ 0.9} & \underline{85.8 $\pm$ 0.5} \\
                  OpenUS         & US-308K       & \textbf{90.3 $\pm$ 0.4} & \textbf{89.4 $\pm$ 0.3} & \textbf{87.8 $\pm$ 0.4} & \textbf{86.3 $\pm$ 0.3} \\
      \bottomrule
    \end{tabular}
    \label{tab1}
  \end{minipage}\hspace{0.002\textwidth}
  \begin{minipage}[t]{0.485\textwidth}
    \centering
    \captionof{table}{Quantitative results on segmentation tasks.}
    \scriptsize
    \setlength{\tabcolsep}{2.5pt}
    \begin{tabular}{ c c c c c c}
      \toprule
      Model           & Pretraining         & \multicolumn{2}{c}{BUS-BRA}           & \multicolumn{2}{c}{TN3K}       \\
      \cmidrule(lr){3-4} \cmidrule(lr){5-6}
                                &                     & DSC(\%)        & IoU(\%)        & DSC(\%)        & IoU(\%)        \\
      \midrule
      \multicolumn{6}{c}{Supervised} \\
      \midrule
                  U-Net           & ImageNet            & 87.0 $\pm$ 1.3 & 79.3 $\pm$ 1.4 & 75.4 $\pm$ 0.6 & 63.3 $\pm$ 0.6  \\
                  U-Net++         & ImageNet            & 86.7 $\pm$ 1.0 & 79.2 $\pm$ 0.8 & 76.7 $\pm$ 0.7 & 66.4 $\pm$ 0.4  \\
                  nnUNet          & /                   & 89.8 $\pm$ 1.6 & 81.2 $\pm$ 1.3 & 80.4 $\pm$ 0.9 & 71.4 $\pm$ 0.9  \\
                  SegFormer       & ImageNet            & 85.4 $\pm$ 1.9 & 74.7 $\pm$ 1.6 & 78.9 $\pm$ 0.9 & 68.6 $\pm$ 1.1  \\
                  UMamba          & /                   & 89.9 $\pm$ 1.7 & 81.3 $\pm$ 1.5 & 80.2 $\pm$ 1.0 & 70.6 $\pm$ 1.0  \\
      \midrule
      \multicolumn{6}{c}{SSL/FM} \\
      \midrule
                  SimMIM          & ImageNet            & 88.1 $\pm$ 0.9 & 80.4 $\pm$ 1.0 & 80.3 $\pm$ 1.2 & 70.5 $\pm$ 1.8  \\
                  MedSAM          & MedImage-1.57M      & 85.6 $\pm$ 1.1 & 74.7 $\pm$ 1.2 & 72.8 $\pm$ 2.0 & 62.2 $\pm$ 1.5  \\
                  DeblurringMAE   & US-Esaote-280K      & \underline{88.8 $\pm$ 1.3} & \underline{81.3 $\pm$ 1.1} & \textbf{82.9 $\pm$ 2.5} & \textbf{73.8 $\pm$ 1.0}  \\
                  USFM            & 3M-US-2M            & 84.5 $\pm$ 2.2 & 74.8 $\pm$ 2.5 & 74.8 $\pm$ 2.1 & 64.1 $\pm$ 1.9  \\
                  OpenUS         & US-308K              & \textbf{91.0 $\pm$ 0.9} & \textbf{83.5 $\pm$ 1.0} & \underline{82.7 $\pm$ 1.2} & \underline{73.1 $\pm$ 1.1}  \\
      \bottomrule
    \end{tabular}
    \label{tab2}
  \end{minipage}
  
\end{figure*}

\section{Experiments}
\subsection{Datasets and Experimental Settings}
We conducted experiments on 42 publicly available ultrasound (US) image datasets, which have a total of 308,584 images covering 12 human organs. The first 38 datasets are utilized for pre-training, where we sample $G=2$ global views and $L=8$ local views, with image sizes set to $224 \times 224$ and $96 \times 96$, respectively. We take VMamba-S \cite{49liu2024vmamba} as the backbone for both the teacher and student networks, and perform 150 epochs of pre-training. In downstream tasks, we perform classification on the BUSI \cite{55al2020dataset} (breast cancer) and Fetal Planes \cite{56burgos2020evaluation} datasets, and segmentation on the BUSBRA \cite{58gomez2024bus} (breast lesions) and TN3K \cite{57gong2023thyroid} (thyroid nodule) datasets. Our implementation is based on iBOT, and more experimental details can be found in \textit{Supplementary Material}.

\subsection{Comparison with Prior Work}
We compare our method with the recent state-of-the-art (SOTA) US image pre-training model, USFM \cite{b5}, the first pre-training foundation model (FM) developed on two million \emph{private} US images spanning multiple organs and devices. We also report results from other pre-training methods, including DeblurringMAE \cite{b44}, which introduces a deblurring-based MIM strategy specifically designed for thyroid nodule ultrasound pre-training, and MedSAM \cite{66ma2024segment}, which is developed using a large-scale medical image dataset comprising 1.5 million image-mask pairs. We also compare against recent SOTA SSL frameworks, including SimMIM \cite{b21}, DINO \cite{51caron2021emerging}, DINOv2 \cite{59oquab2024dinov}, and iBOT \cite{b38}. Additionally, we compare the latest supervised learning methods for natural or medical image analysis, including U-Net \cite{60ronneberger2015u}, U-Net++ \cite{61unet++}, nnUNet \cite{62isensee2021nnu}, SegFormer \cite{63xie2021segformer}, UMamba \cite{64ma2024u}, VMamba \cite{49liu2024vmamba}, ResNet50 \cite{65he2016deep}, and ViT \cite{51caron2021emerging}, for downstream segmentation and classification tasks. 

\subsubsection{Quantitative Evaluation.} 
We observe that our method outperforms existing SOTA methods, as shown in Tab.~\ref{tab1} and \ref{tab2}. Particularly, compared to the FM-based USFM, our \textit{OpenUS} achieves improvements of $7.9\%$ DSC and $9.0\%$ IoU on the TN3K segmentation task, as well as $6.5\%$ DSC and $8.7\%$ IoU on the BUS-BRA segmentation task. Additionally, compared to the FM-based DeblurringMAE pretrained on 280K US images of the same modality as TN3K, our \textit{OpenUS} achieves comparable performance, with only $0.2\%$ lower DSC and $0.7\%$ lower IoU on the TN3K segmentation task. Moreover, \textit{OpenUS} attains notable improvements of $2.2\%$ in DSC and $2.2\%$ in IoU on the BUS-BRA segmentation task, $2.1\%$ in ACC and $3.5\%$ in F1-score on the Fetal Planes classification task, and $2.9\%$ in ACC and $3.9\%$ in F1-score on the BUSI classification task. These improvements can be attributed to our proposed pre-training approach, which integrates self-adaptive masking and global-local-view mask modeling with contrastive learning. Clearly, compared to other FM/SSL-based methods such as SimMIM and MedSAM, our \textit{OpenUS} demonstrates substantial advantages in downstream segmentation tasks. Although performance gains on the classification task are modest compared to the latest USFM, our \textit{OpenUS} achieves substantial progress using significantly fewer pre-training ultrasound images (296K vs. 2M), while still obtaining slightly better results on the Fetal Planes (by $0.1\%$ ACC and $0.3\%$ F1) and BUSI (by $0.3\%$ ACC and $0.5\%$ F1) classification tasks.

\subsubsection{Qualitative Evalutation.} 
\begin{figure*}[t]
\centering
\includegraphics[width=1\textwidth]{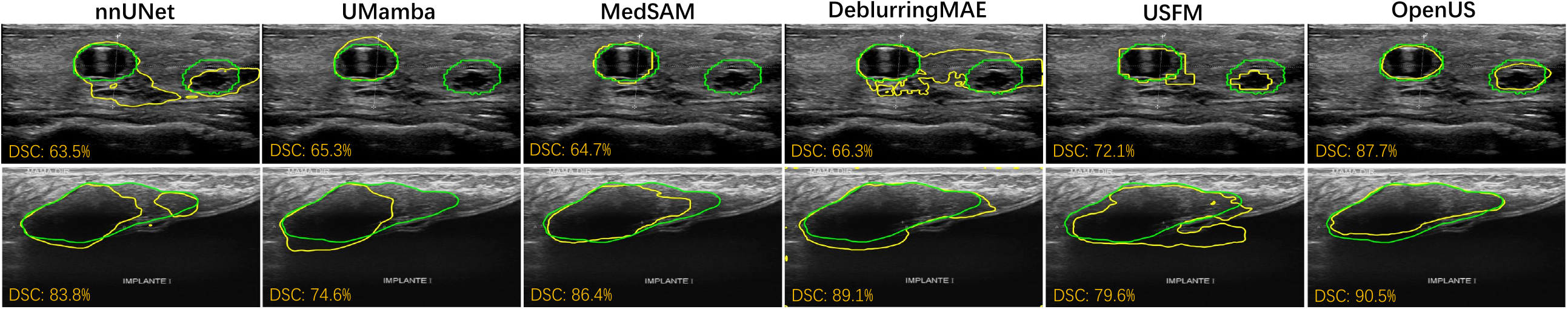} 
\caption{Visualization of US segmentation results on TN3K and BUS-BRA. The ground truth is depicted in green, and the prediction is shown in yellow.}
\label{fig3}
\end{figure*}

We visualize segmentation results in Fig.~\ref{fig3}. Clearly, \textit{OpenUS} outperforms other methods in accurately recognizing hazy tissue borders and demonstrates greater robustness to speckle in US images. Specifically, \textit{OpenUS} demonstrates superior visual results in segmenting both large and small thyroid nodule regions ($1^{st}$ rows in Fig.~\ref{fig3}). We attribute these improvements to our self-adaptive masking strategy, which effectively captures more precise and clinically relevant details from US images. Similarly, \textit{OpenUS} also exhibits strong performance in segmenting breast tumour ($2^{nd}$ row in Fig.~\ref{fig3}).

\subsubsection{Label Efficiency Analysis.}
\begin{figure}[t]
\centering
\includegraphics[width=0.48\textwidth]{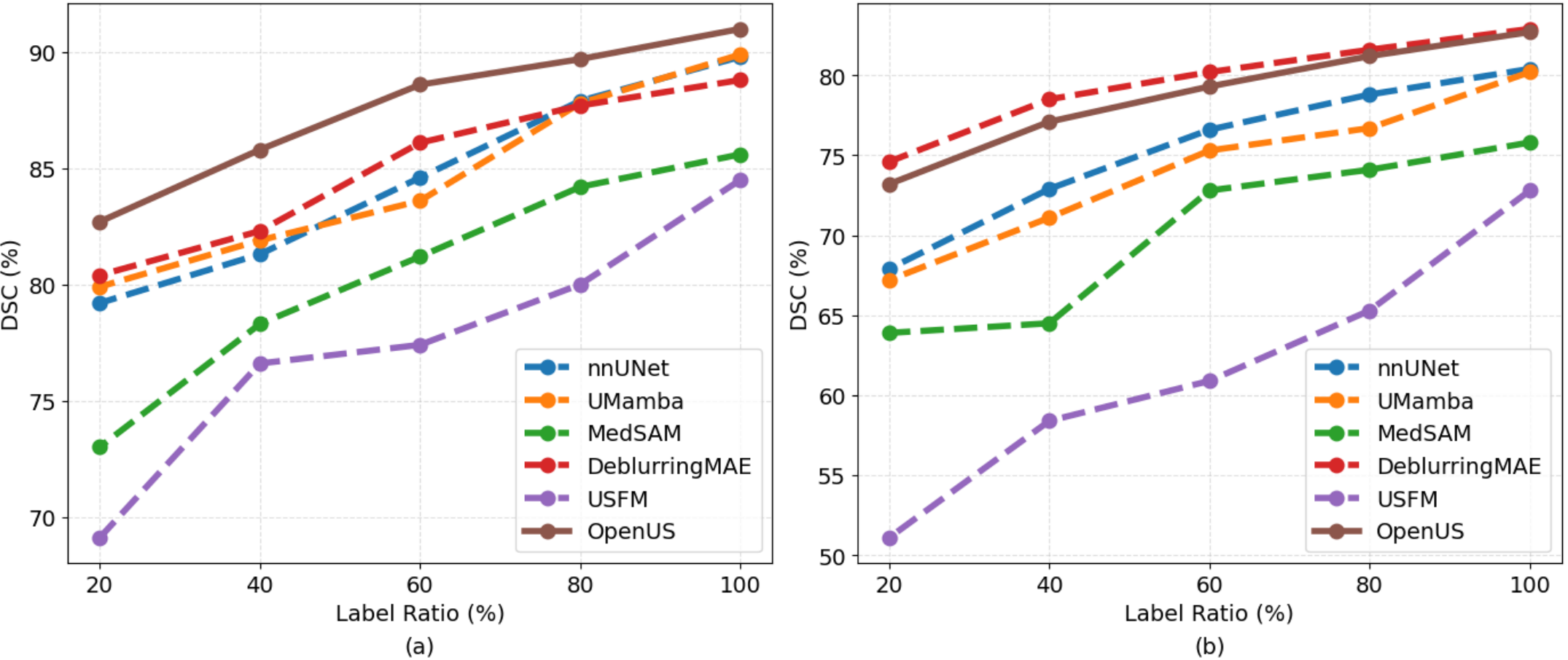} 
\caption{Label efficiency experiments of the downstream segmentation tasks: (a) BUS-BRA and (b) TN3K. The DSC scores (\%) of the model trained at different label ratios are reported.}
\label{fig4}
\end{figure}

The label efficiency results for the downstream segmentation tasks are presented in Fig.~\ref{fig4}. When developing downstream tasks on training sets with varying label ratios, \textit{OpenUS} consistently outperformed both supervised and most SSL methods in terms of label efficiency. Even with only a $20\%$ label available, \textit{OpenUS} achieves satisfactory performance, with DSC of $73.2\%$ and $82.7\%$ for segmentation downstream tasks of thyroid nodule and breast tumour. With the label ratio increasing from $20\%$ to $40\%$, \textit{OpenUS} still demonstrates robust performance, achieving DSC of $77.1\%$ and $85.8\%$ in the thyroid nodule and breast tumor segmentation. When the label ratio increased to $60\%$, \textit{OpenUS} achieves segmentation performance on thyroid and breast comparable to that with a $100\%$ label ratio. Compared to \textit{OpenUS}, DeblurringMAE, pre-trained on 280K thyroid-only US images, achieved a 0.2\%-1.4\% higher DSC score across all label ratios on the TN3K segmentation task, likely due to its enhanced feature extraction capability for thyroid nodule images across all label ratios. However, for breast tumor segmentation, DeblurringMAE consistently underperformed relative to \textit{OpenUS} across all label ratios. More results for label efficiency in the classification task can be found in \textit{Supplementary Material}.

\subsection{Ablation Studies}
\subsubsection{Global-Local-View Mask Strategies.}
Tab.~\ref{tab3} presents the impact of global-view masking and combined global-local-view masking strategies. All experiments use the same masking ratio. For the single-view, different masking strategies are applied to global views. We evaluate our approach against two baseline strategies. The random blockwise mask strategy \cite{b19} applies masking uniformly across the spatial domain, while the attention mask selects tokens to mask based on attention maps from global views. In contrast, our proposed self-adaptive mask employs a teacher-student feedback mechanism. This mechanism calculates an $ALP$ score from global views to guide the masking process, significantly outperforming these baselines with 82.1\% on segmentation and 90.2\% on classification tasks, respectively. In the multi-view setting, applying a self-adaptive mask to global views and a random blockwise mask to local views yields superior performance on both classification (90.3\% in ACC) and segmentation (82.7\% in DSC) tasks. These complementary strategies enable the model to learn more comprehensive details from the US features.
\begin{table}[ht]
\centering
\caption{Comparison of mask strategies on different views. RBW - Random Blockwise.}
\small
\setlength{\tabcolsep}{4.5pt}
\begin{tabular}{cccccc}
\toprule
\multirow{2}{*}{Views} & \multicolumn{2}{c}{Mask Strategies} &  Cla.& Seg. \\
\cmidrule(lr){2-3} 
& Global & Local  & ACC(\%)$\boldsymbol{\uparrow}$ & DSC(\%)$\boldsymbol{\uparrow}$ \\
\midrule
\multirow{2}{*}{Single-view} 
    & RBW       & $\usym{2717}$   & 88.6 $\pm$ 0.2 & 79.1 $\pm$ 1.1 \\
    & Attention    & $\usym{2717}$   & 89.1 $\pm$ 0.4 & 80.9 $\pm$ 1.0 \\
    & Self-adaptive& $\usym{2717}$   & 90.2 $\pm$ 0.6 & 81.6 $\pm$ 1.3 \\
\midrule
\multirow{2}{*}{Multi-view} 
    & RBW       & RBW  & 89.4 $\pm$ 0.3 & 81.2 $\pm$ 1.4 \\
    & Attention    & RBW  & 89.7 $\pm$ 0.5 & 81.7 $\pm$ 1.1 \\
    & Self-adaptive& RBW  & \textbf{90.3 $\pm$ 0.4} & \textbf{82.7 $\pm$ 1.2} \\
\bottomrule
\end{tabular}
\label{tab3}
\end{table}

\subsubsection{Analysis of Components.} We conducted an ablation study to analyse the framework's components using two downstream tasks: Fetal Planes and TN3K.  As demonstrated from the first row in Tab.~\ref{tab4}, the results indicate that the contrastive learning framework performs reasonably well on classification but lacks robustness in pixel-level tasks, particularly segmentation. Specifically, combining contrastive learning with global and local-view MIM boosted segmentation performance, increasing the DSC score from 79.1\% to 82.7\%. Additionally, we observed that the global-view MIM was superior to the local-view MIM on the segmentation task, resulting in a 1.8\% increase in DSC. These results highlight the complementary strengths of the two methods: the powerful instance discrimination from contrastive learning and the robust, global and local pixel-level feature acquisition from MIM. 
\begin{table}[ht]
\centering
\caption{Ablation study on components analysis.} 
\small
\begin{tabular}{ccccc}
\toprule
\multirow{2}{*}{CL} & \multicolumn{2}{c}{MIM} & Cla. & Seg. \\
\cmidrule(lr){2-3}
& Global & Local&  ACC(\%)& DSC(\%) \\
\midrule
$\usym{2713}$  & $\usym{2717}$ & $\usym{2717}$ & 89.7 $\pm$ 0.3 & 79.1 $\pm$ 1.1\\
$\usym{2713}$ & $\usym{2713}$ & $\usym{2717}$ & 90.2 $\pm$ 0.6 & 81.6 $\pm$ 1.3 \\
$\usym{2713}$ & $\usym{2717}$ & $\usym{2713}$ & 89.9 $\pm$ 0.5 & 80.3 $\pm$ 1.4 \\
$\usym{2713}$ & $\usym{2713}$ & $\usym{2713}$ & \textbf{90.3 $\pm$ 0.4 }& \textbf{82.7 $\pm$ 1.2} \\
\bottomrule
\end{tabular}
\label{tab4}
\end{table}

\subsubsection{Comparison with Other SSL Methods.} We evaluated the \textit{OpenUS} model against SOTA SSL methods on a downstream breast tumor classification task. To ensure a fair comparison, all frameworks utilized the same VMamba teacher and student networks, were pre-trained on all 38 pre-trained US datasets, and were then tested on the held-out BUSI dataset. As shown in Tab.~\ref{tab5}, \textit{OpenUS} outperforms the other SSL frameworks under identical experimental settings. 

\begin{table}[ht]
\centering
\caption{Comparison of different frameworks on the BUSI downstream classification task.}
\small
\begin{tabular}{c c c c}
\toprule
Framework & Teacher/Student & ACC(\%)$\boldsymbol{\uparrow}$ & F1 (\%)$\boldsymbol{\uparrow}$ \\
\midrule
DINO      & VMamba & 84.3 $\pm$ 0.5 & 83.6 $\pm$ 0.6 \\
iBOT      & VMamba & 85.6 $\pm$ 0.4 & 84.3 $\pm$ 0.5 \\
DINOv2    & VMamba & 85.9 $\pm$ 0.3 & 84.5 $\pm$ 0.4 \\
OpenUS    & VMamba & \textbf{87.8 $\pm$ 0.3} & \textbf{86.3 $\pm$ 0.5} \\
\bottomrule
\end{tabular}
\label{tab5}
\end{table}
\section{Conclusion}
In this study, we developed a universal US image foundation model named \textit{OpenUS}, characterized by label efficiency and downstream task adaptability. 
We propose a self-adaptive masking strategy to capture the clinically relevant and generalizable US features for global and local views. 
Importantly, \textit{OpenUS} was developed using 42 publicly available US datasets to promote open access, reproducibility and ensure wide applicability. 
{
    \small
    \bibliographystyle{ieeenat_fullname}
    \bibliography{main}
}

\clearpage
\setcounter{page}{1}
\maketitlesupplementary
This supplementary material begins by detailing the pre-training datasets in Sec.~\ref{SecA} and providing model implementation details for reproducibility in Sec.~\ref{SecB}. Sec.~\ref{SecC} reports additional results on label efficiency, followed by further visualizations and qualitative analysis of segmentation results in Sec.~\ref{SecD}. Finally, we include additional ablation experiments in Sec.~\ref{SecE} and a discussion of the study's limitations, future work, and broader impacts in Sec.~\ref{SecF}.

\section{Details of Pre-training Datasets.}
\label{SecA}
Due to its significant domain gap from other medical imaging modalities, developing a foundation model for ultrasound (US) Image Analysis requires a large-scale, specialized dataset. To build the fully open-access dataset, we first followed US-43d \cite{sm1_meyer2024ultrasam}, then we sourced US data from various online platforms: Kaggle, Mendeley dataset, Zenodo, Google Scholar, GitHub and ResearchGate. The resulting collection comprises 38 datasets for pre-training, encompassing 10 different organs, multiple clinical centers, and various device manufacturers, as shown in Tab. \ref{tab:us_datasets}.

\begin{table*}[ht]
\centering
\caption{Overview of pre-training ultrasound datasets.}
\small
\begin{tabular}{lcc}
\toprule
Dataset & Organ & Number \\
\midrule
105US \cite{sm2_105_hann2017algorithm} & Liver & 105 \\
AbdomenUS \cite{sm3_abdomenUS_vitale2020improving} & Abdomen & 617 \\
ACOUSLIC \cite{sm4_acouslic_gleed2023towards}& Abdomen & 6620 \\
ASUS \cite{sm5_asus_ungi2020automatic} & Abdomen & 2865 \\
AUL \cite{sm6_aul_xu2023improving}& Liver & 735 \\
brachial plexus \cite{sm7_brachial_plexus_tyagi2024nerve}& Nerve & 40788 \\
BrEaST \cite{sm8_breast_pawlowska2024curated}& Breast & 252 \\
BUID \cite{sm9_buid_ardakani2023open}& Breast & 232 \\
BUS\_UC \cite{sm10_bus_us_iqbal2024memory}& Breast & 810 \\
BUS\_UCML \cite{sm11_bus_uclm_vallez2025bus}& Breast & 264 \\
BUS \cite{sm12_bus_yap2017automated} & Breast & 163 \\
CAMUS \cite{sm13_camus_leclerc2019deep} & Cardiac & 19232 \\
CCAUI \cite{sm14_ccaui_momot2022common} & Echocardiogram & 1100 \\
DDTI \cite{sm15_ddti_tg3k_tn3k_pedraza2015open}& Thyroid & 637 \\
EchoNet-Dynamic \cite{sm16_echonet_dynamic_ouyang2020video} & Echocardiogram & 20048 \\
EchoNet-Pediatric \cite{sm17_echonet_pediatric_reddy2023video} & Echocardiogram & 15450 \\
FALLMUD \cite{sm18_fallmud_michard2021aw} & Muscle & 813 \\
FASS \cite{sm19_fass_da2023fetal} & Fetal abdomen & 1588 \\
Fast-U-Net \cite{sm20_fast_unet_ashkani2022fast} & Abdominal Circumference  and Head Circumference & 1411 \\
GIST514-DB \cite{sm21_gist514_he2022query2} & Gastrointestinal Stromal Tumour & 43656 \\
HC \cite{sm22_kineyus_singla2023open} & Fetal head & 1334 \\
kidneyUS & Kidney & 487 \\
LUSS\_phantom \cite{sm23_luss_phantom_mclaughlan2024lung} & Lung & 564 \\
MicroSeg \cite{sm24_microseg_jiang2024microsegnet} & Prostate & 1931 \\
MMOTU-2D \cite{sm25_mmotu2d_3d_zhao2022mmotu} & Ovarian Tumor & 1469 \\
MMOTU-3D \cite{sm25_mmotu2d_3d_zhao2022mmotu} & Ovarian Tumor & 170 \\
regPro \cite{sm26_regpro_hamid2019smarttarget} & Prostate & 4706 \\
S1 \cite{sm27_s1_Guo2021_expandedUNet}& Breast & 201 \\
Segthy \cite{sm28_segthy_kronke2022tracked} & Thyroid & 12737 \\
STMUS\_NDA \cite{sm29_stmus_nda_marzola2021deep} & Transverse musculoskeletal & 4355 \\
STU-Hospital \cite{sm34_stu_hospital_zhuang2019rdau} & Breast & 42 \\
TG3K \cite{sm15_ddti_tg3k_tn3k_pedraza2015open} & Thyroid & 3585 \\
Thyroid US Cineclip \cite{sm30_stanford_aimi_thyroid_cineclip_2021} & Thyroid & 17412 \\
UPBD \cite{sm31_ubpd_ding2022mallesnet} & Brachial Plexus & 955 \\
US nerve Segmentation \cite{sm32_ultrasound_nerve_segmentation_kaggle_montoya2016ultrasound} & Nerve & 11134 \\
USAnotAI \cite{sm33_usanotai_git2020usanotai} & Anatomical regions & 366 \\
Cactus \cite{sm35_cactus_elmekki2025cactus}& Cardiac & 37736 \\
IUGC25 \cite{sm37_iugc25_bai2025landmark}& Fetal head & 33466 \\

\bottomrule
\end{tabular}
\label{tab:us_datasets}
\end{table*}

\section{Implementation Details.}
\label{SecB}
\subsection{Pre-training Settings.} The pre-training dataset is augmented using techniques such as random horizontal flipping, color jitter, Gaussian blur, and exposure adjustment. The decoder consists of a lightweight, one-layer head \cite{b21}, and the model is trained with an $\ell_2$ loss. The proposed model contains 59M parameters. Pre-training was performed on four NVIDIA GH200 GPUs, with detailed training configurations provided in Tab.~\ref{tab:settings}. 

\begin{table}[ht]
\centering
\caption{Pre-training settings.}
\small
\begin{tabular}{l|l}
\toprule
Config & Value \\
\midrule
optimizer                & AdamW \cite{sm38_loshchilov2018decoupled} \\
optimizer momentum       & $\beta_1, \beta_2 = 0.9,\, 0.999$ \\
weight decay             & $4\mathrm{e}{-2}$ \\
base learning rate       & $5\mathrm{e}{-4}$ \\
learning rate schedule   & cosine \cite{sm39_loshchilov2016sgdr} \\
warmup epochs            & 30 \\
pretraining epochs       & 150 \\
batch size               & 128 \\
temperature parameters   & $\tau_t,\, \tau_s = 0.04,\, 0.07$ \\
masking rate                & $r = 0.8$ \\
momentum coefficient     & $\lambda = 0.996$ \\
$ALP$ coefficient        & $\alpha \in [0.1,0.9]$ \\
$ALP$ schedule           & cosine schedule \\
\bottomrule
\end{tabular}
\label{tab:settings}
\end{table}

\subsection{Evaluation Methodology.} For downstream fine-tuning, the procedure is as follows: (1) \textbf{Classification}: Our analysis was performed on two US datasets: the BUSI dataset \cite{55al2020dataset}, with 1,560 images for 3-class classification (benign, malignant, and normal), and the Fetal Planes dataset \cite{56burgos2020evaluation}, with 12,400 images for 6-class fetal plane classification (Abdomen, Brain, Femur, Thorax, maternal cervix, and other). The BUSI and Fetal Planes datasets were partitioned into training, validation, and testing sets, comprising 1090/230/240 and 7440/2480/2480 US images, respectively. All input US images were resized to $224\times224$. The network architecture consists of a pre-trained backbone and a new, randomly initialized linear classification head, which was subsequently trained for 100 epochs. For optimization, we employed the Stochastic Gradient Descent (SGD) with a batch size of 16, a learning rate of 1e-3, and a momentum value of 0.9. Additionally, the primary evaluation metrics for classification performance were Accuracy (ACC) and the F1-score. (2) \textbf{Segmentation}: This evaluation utilizes two US datasets: the BUSBRA \cite{58gomez2024bus} (breast lesions) and TN3K \cite{57gong2023thyroid} (thyroid nodule) datasets. The BUSBRA dataset is partitioned into 1,200 training, 299 validation, and 376 testing images, while the TN3K dataset consists of 2,303 training, 576 validation, and 614 testing images. Each US image is annotated with a ground truth mask that delineates the boundaries of the polyp region. We employ the channel-aware Mamba decoder as the segmentation decoder. We optimized the network using the AdamW optimizer, setting the learning rate to 1e-4, weight decay to 5e-2, and the batch size to 16. All input US images are resized to $224\times224$ and fine-tuned for 100 epochs. Furthermore, the primary evaluation metrics for segmentation performance were Dice Similarity Coefficient (DSC) and Intersection over Union (IoU).

\begin{figure}[t]
\centering
\includegraphics[width=0.48\textwidth]{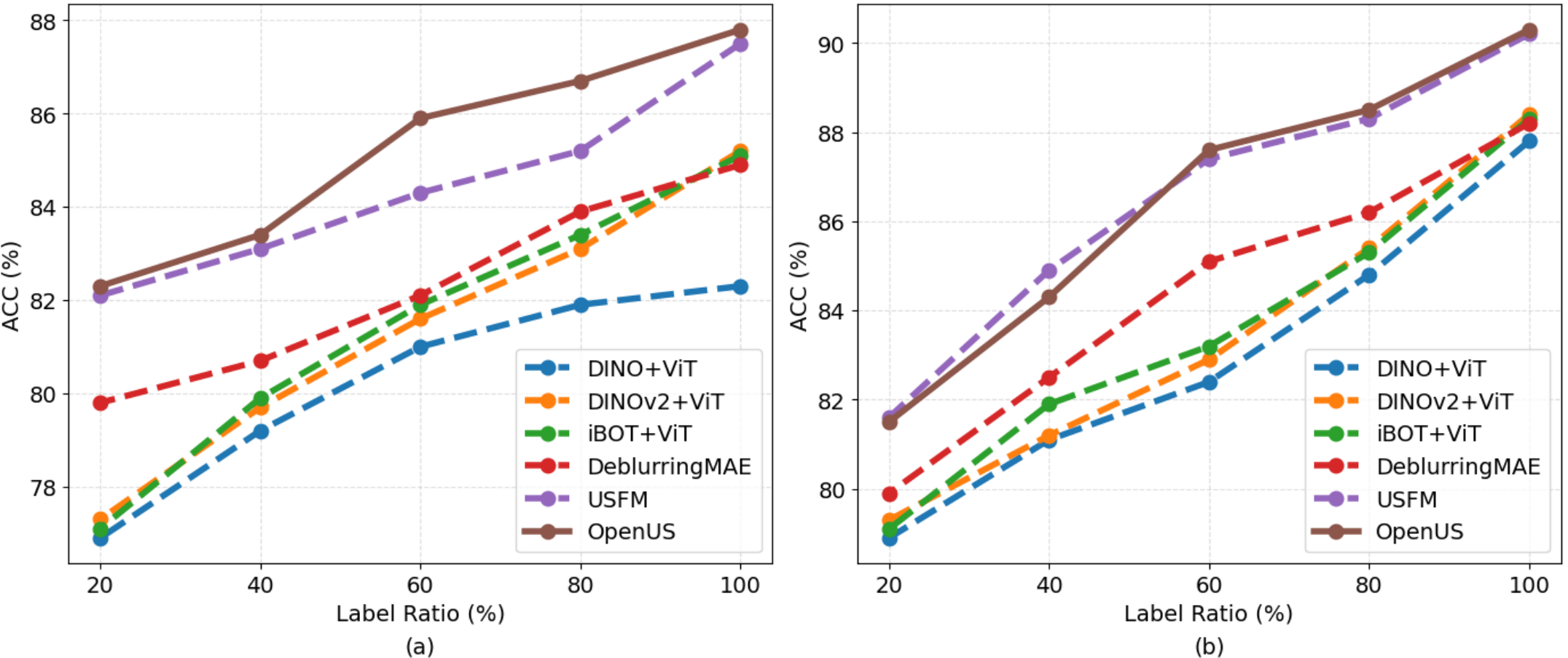} 
\caption{Label efficiency experiments of the downstream classifications tasks: (a) BUSI and (b) Fetal Planes. The ACC values (\%) of the model trained at different label ratios are reported.}
\label{fig5_sm}
\end{figure}

\begin{figure}[t]
\centering
\includegraphics[width=0.48\textwidth]{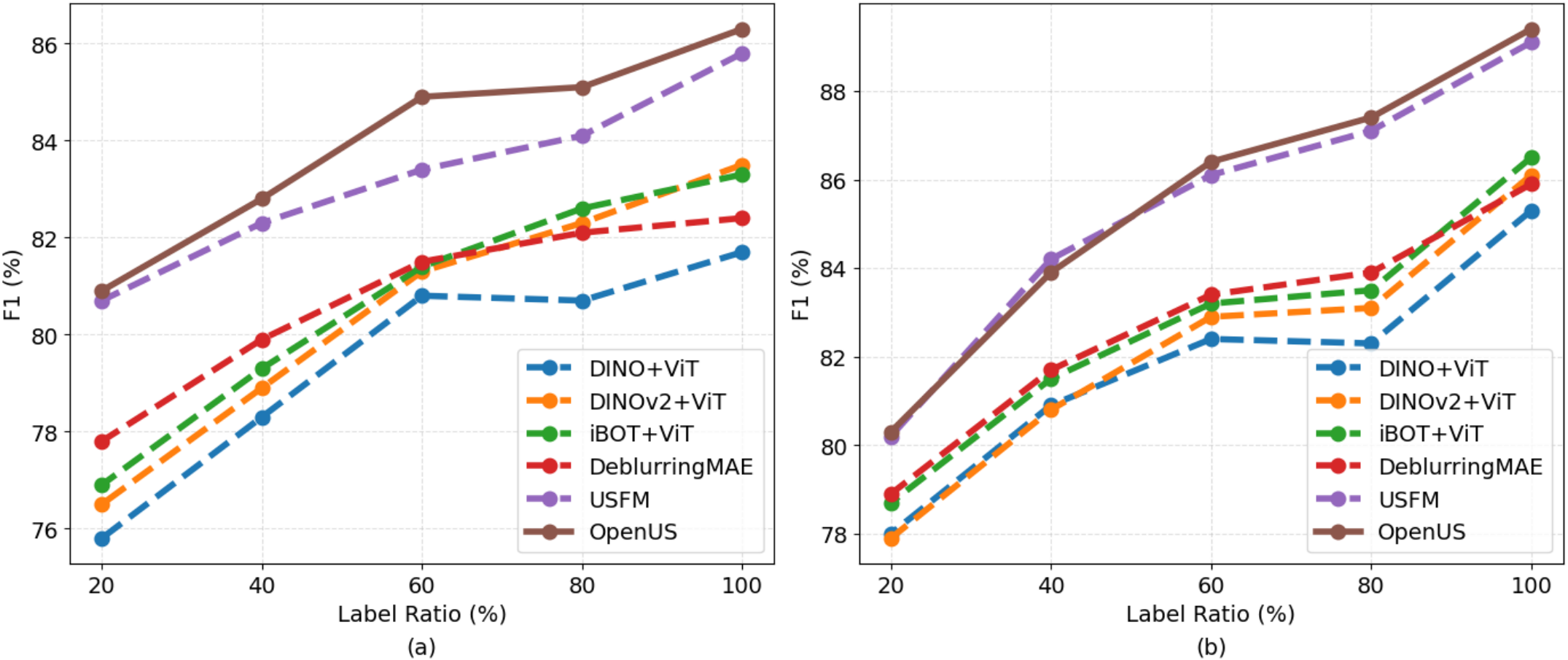} 
\caption{Label efficiency experiments of the downstream classifications tasks: (a) BUSI and (b) Fetal Planes. The F1 scores (\%) of the model trained at different label ratios are reported.}
\label{fig6_sm}
\end{figure}

\begin{figure}[t]
\centering
\includegraphics[width=0.48\textwidth]{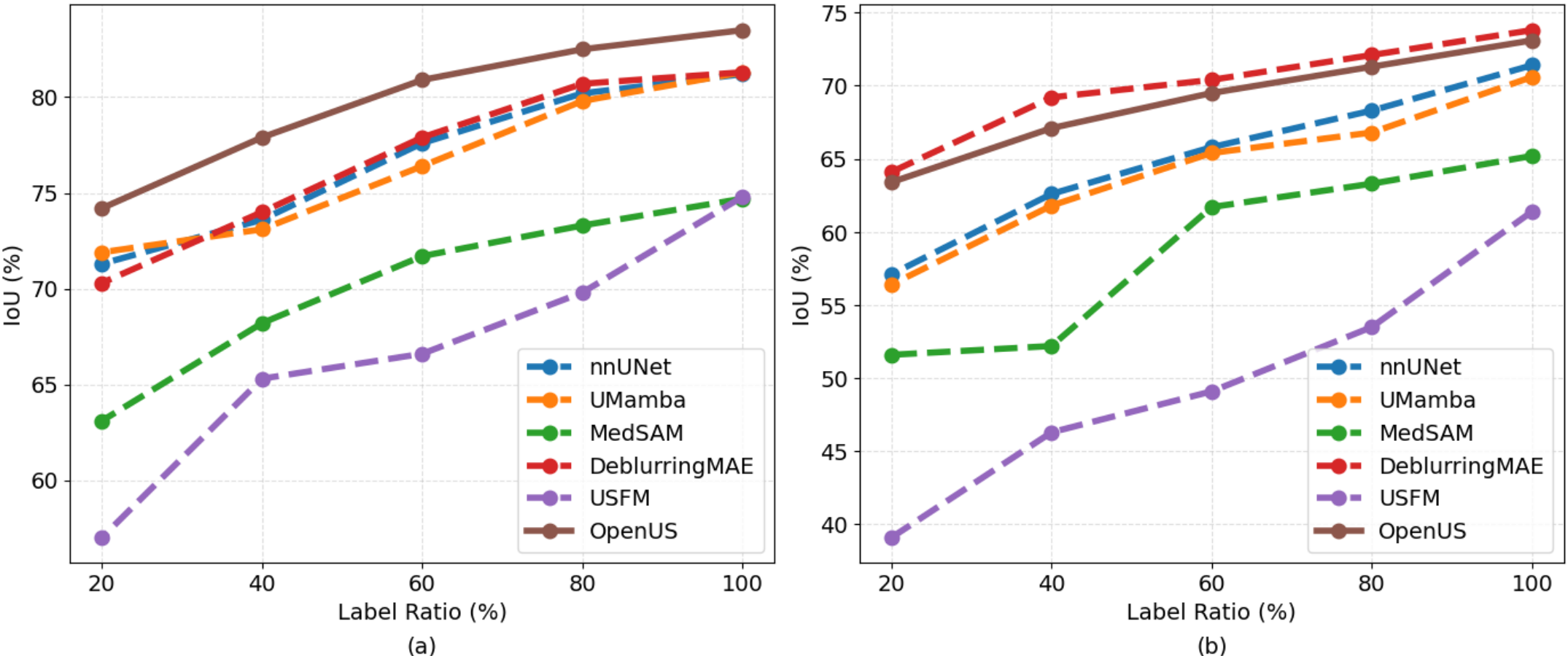} 
\caption{Label efficiency experiments of the downstream segmentation tasks: (a) BUSBRA and (b) TN3K. The IoU (\%) of the model trained at different label ratios are reported.}
\label{fig8_sm}
\end{figure}

\begin{figure*}[t]
\centering
\includegraphics[width=1\textwidth]{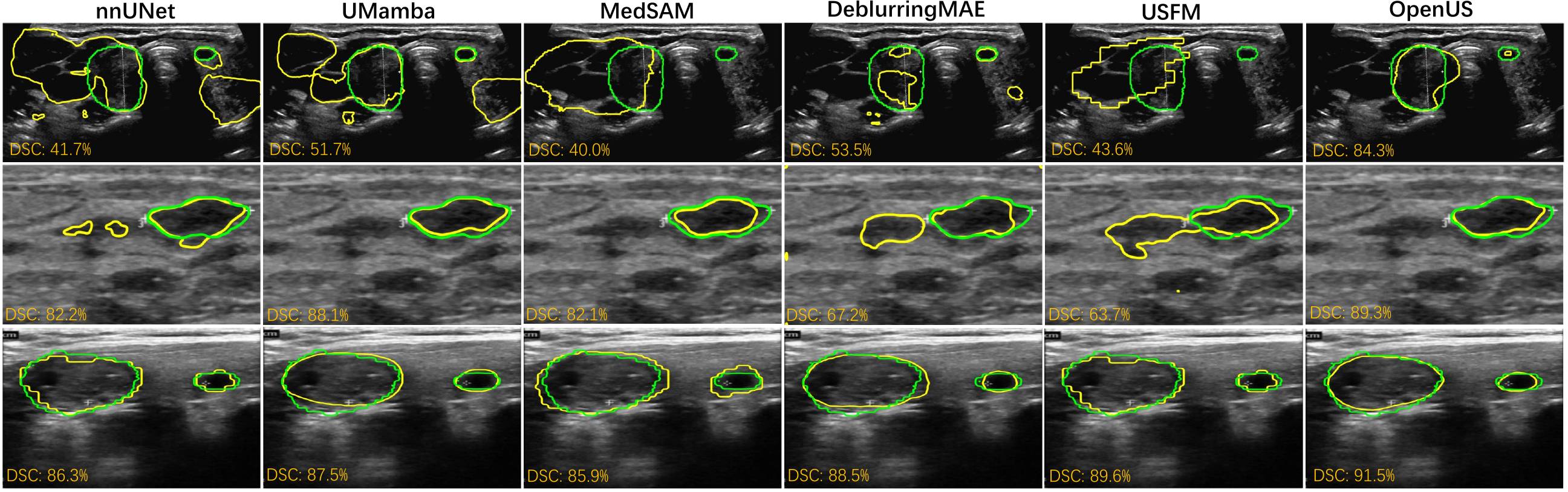} 
\caption{Visualization of US segmentation results on TN3K and BUS-BRA. The ground truth is depicted in green, and the prediction is shown in yellow.}
\label{fig7_sm}
\end{figure*}

\section{Additional Label Efficiency Analysis.} \label{SecC}
The label efficiency results for the downstream classification tasks are presented in Fig. \ref{fig5_sm} and Fig. \ref{fig6_sm}. In the BUSI classification task, the pre-trained on ImageNet or thyroid nodule ultrasound SSL methods, i.e., DINO, DINOv2, iBOT, and DeblurringMAE, showed strong dependence on the amount of annotated data. At a 20\% label ratio, these four methods failed to achieve a satisfactory classification performance, with ACC values and F1 scores below 80\%. In the Fetal Planes classification task, all methods besides \textit{OpenUS} and USFM demonstrated suboptimal performance when using a low ratio of labeled data. Specifically, at a 20\% label ratio, their ACC values and F1 scores both fell below 80\%. 

Compared to \textit{OpenUS}, USFM pre-trained on 2 million private US images shows improved performance over our \textit{OpenUS} for Fetal Planes classification task, particularly with the lower label ratio (20\% and 40\%).  Conversely, at higher label ratios (60\%, 80\%, and 100\%), \textit{OpenUS} surpasses the performance of USFM, achieving ACC of $90.3\%$, $88.5\%$ and $87.6\%$, and F1 of $89.4\%$, $87.5\%$ and $86.3\%$. This result indicates that \textit{OpenUS} effectively learns features from US images across various organs and possesses strong generalizability, even though it was pre-trained on a much smaller US dataset than USFM. 

The additional label efficiency results for the downstream segmentation tasks are demonstrated in Fig.~\ref{fig8_sm}. \textit{OpenUS} exhibits remarkable label efficiency, consistently outperforming both supervised and the majority of semi-supervised learning (SSL) methods on the BUSBRA and TN3K datasets. Even with a highly limited training set of only $20\%$ of labels, the model attained impressive IoU scores of $74.2\%$ in breast tumor segmentation and $63.4\%$ in thyroid nodule segmentation. The performance of OpenUS scales effectively with the amount of labeled data. Increasing the label ratio to $40\%$ boosts the IoU scores to $77.9\%$ for breast tumor and $67.1\%$ for thyroid nodule segmentation. Remarkably, at just a $80\%$ label ratio, the model's performance becomes comparable to that of a fully supervised model using $100\%$ of the labels. Although DeblurringMAE exhibits slightly better performance in thyroid nodule segmentation, achieving a 0.7\% to 2.1\% higher IoU across all label ratios—this advantage is likely attributable to its pre-training on the same imaging modality as the thyroid nodule task. Conversely, for the breast tumor segmentation task, \textit{OpenUS} demonstrated superior performance to DeblurringMAE at all levels of supervision.

\section{Additional Qualitative Evalutation.} \label{SecD}
Fig.~\ref{fig7_sm} demonstrates the segmentation results of our method and other superivised and self-supervised pre-training methods on the TN3K and BUS-BRA datasets. The results indicate that \textit{OpenUS} outperforms other methods in the accurate recognition of indistinct tissue borders and demonstrates greater robustness to speckle noise in US images. Specifically, for thyroid nodule segmentation tasks involving noisy, low-quality US images ($1^{st}$ and $3^{rd}$ row in Fig.~\ref{fig7_sm}), nnUNet, UMamba, MedSAM, DeblurringMAE, and USFM inaccurately segment artefacts, whereas \textit{OpenUS} remains robust. While some small thyroid nodules occupy very few pixels and are prone to omission ($1^{st}$ row in Fig.~\ref{fig7_sm}), \textit{OpenUS} successfully segments these challenging structures. In the breast cancer segmentation task ($2^{nd}$ row, Fig.~\ref{fig7_sm}), our method produces segmentation masks with smoother and more continuous edges. In contrast, other methods often yield results corrupted by speckle noise or containing inaccurate lesion boundaries.

\section{Additional Ablation Studies} \label{SecE}

\subsection{Masking Ratio.} The impact of different masking ratios is illustrated in Tab.~\ref{tab:mask_ratio_results}. An increase in the masking ratio from 60\% to 80\% yields a notable improvement in performance across two downstream tasks. When the masking ratio in US images reaches 80\%, the task becomes challenging as the model must learn feature correspondences from a very limited set of visible patches. This forces the model to develop more robust and meaningful representations, which in turn boosts its overall learning capacity. In contrast, the model's performance degrades significantly with a 95\% masking ratio, which is the lowest-performing of the four ratios. This indicates that at such a high level of occlusion, the model becomes prohibitively difficult. The extreme scarcity of visible patches prevents the model from effectively learning feature correspondences, leading to a decline in the quality of its learned representations.

\begin{table}[ht]
\centering
\caption{Comparison of classification and segmentation performance for different self-adaptive mask ratios.}
\small
\begin{tabular}{cccc}
\toprule
\multicolumn{2}{c}{Self-Adaptive} & Cla. & Seg. \\
\cmidrule(lr){1-2}
Mask Ratio & & ACC(\%) & DSC(\%) \\
\midrule
60\% & & $89.6 \pm 0.2$ & $81.8 \pm 1.4$ \\
70\% & & $89.5 \pm 0.3$ & $81.9 \pm 1.1$ \\
80\% & & $\mathbf{90.3 \pm 0.4}$ & $\mathbf{82.7 \pm 1.2}$ \\
90\% & & $89.0 \pm 0.5$ & $80.3 \pm 1.3$ \\
\bottomrule
\end{tabular}
\label{tab:mask_ratio_results}
\end{table}

\section{Limitations, Future Work, and Broader Implications.} \label{SecF}
\subsection{Limitations.} We introduce \textit{OpenUS}, a fully open-access foundation model for US image analysis, designed for a novel self-adaptive masked contrastive learning framework. Our \textit{OpenUS} foundation model targets fully open-access, reproducibility, and wide applicability. However, we built \textit{OpenUS} on all public datasets for pre-training and downstream task validation. A key challenge to the reproducibility of \textit{OpenUS} is its reliance on public datasets for pre-training and downstream task validation, as their future availability cannot be guaranteed. Additionally, the extensive pre-training required by \textit{OpenUS} is computationally expensive, demanding significant energy consumption and specialized high-performance hardware, such as GPUs. 

\subsection{Future Work.} First, we will expand the pre-training corpus by integrating additional public US datasets. This will aim to enhance the robustness and generalization capabilities of \textit{OpenUS}. Second, we plan to advance the model beyond its current unimodal (image-only) pre-training by incorporating multimodal data, such as US videos and corresponding textual information. This multi-modal approach is expected to enrich the feature representations learned during self-supervision, leading to a more powerful model. Finally, we will demonstrate the utility and versatility of our model by fine-tuning the pre-trained \textit{OpenUS} for a broader range of downstream tasks, such as medical image detection, enhancement, and generation.

\subsection{Broader Implications.} Our approach underscores the potential of SSL for US image analysis. By pre-training on large-scale unlabeled US images, our method mitigates the dependence on costly annotated medical data and improves the label efficiency. The resulting models serve as a versatile foundation for diverse downstream applications, including classification and segmentation, thereby offering a scalable solution to enhance the quality and efficiency of clinical diagnostics.


\end{document}